%% file: main.tex
\documentclass[runningheads]{llncs}

\usepackage{graphicx}
\usepackage{comment}
\usepackage{amsmath,amssymb} 
\usepackage{color}
\usepackage[dvipsnames]{xcolor}
\usepackage{booktabs}
\usepackage{multirow}

\usepackage{epsfig}
\usepackage{calrsfs}
\usepackage{subcaption}
\usepackage{algorithm}
\usepackage{algorithmic}
\usepackage{lipsum}

\let\svthefootnote\thefootnote

\newcommand\blankfootnote[1]{%
  \let\thefootnote\relax\footnotetext{#1}%
  \let\thefootnote\svthefootnote%
}
\let\svfootnote\footnote
\renewcommand\footnote[2][?]{%
  \if\relax#1\relax%
    \blankfootnote{#2}%
  \else%
    \if?#1\svfootnote{#2}\else\svfootnote[#1]{#2}\fi%
  \fi
}

\def\@onedot{\ifx\@let@token.\else.\null\fi\xspace}                
\def\eg{\emph{e.g}\@onedot} \def\Eg{\emph{E.g}\@onedot}             
\def\ie{\emph{i.e}\@onedot} \def\Ie{\emph{I.e}\@onedot}             
\def\cf{\emph{cf}\@onedot} \def\Cf{\emph{Cf}\@onedot}               
\def\etc{\emph{etc}\@onedot} \def\vs{\emph{vs}\@onedot}             
\def\wrt{w.r.t\@onedot} \def\dof{d.o.f\@onedot}                     
\def\etal{\emph{et al}\@onedot}                                    

\begin{document}

\title{Large Scale Holistic Video Understanding}
\authorrunning{Diba, Fayyaz, Sharma et al.}
\author{
    Ali Diba$^{1,5\star}$, Mohsen Fayyaz$^{2,\star}$, Vivek Sharma$^{3,\star}$,  \\ Manohar Paluri,  J\"{u}rgen Gall$^{2}$, Rainer Stiefelhagen$^{3}$, Luc Van Gool$^{1,4,5}$
}
\institute{$^{1}$KU Leuven, $^{2}$University of Bonn,$^{3}$KIT, Karlsruhe,  $^{4}$ETH Z\"{u}rich, $^{5}$Sensifai \\ 
\tt\small \{firstname.lastname\}@kuleuven.be, \{lastname\}@iai.uni-bonn.de, \{firstname.lastname\}@kit.edu, Balamanohar@gmail.com}

\pagestyle{headings}
\mainmatter
\def\ECCVSubNumber{3994}  

\title{Large Scale Holistic Video Understanding} 

\maketitle
\begin{abstract}
\footnote[]{$^{\star}$Ali Diba, Mohsen Fayyaz and Vivek Sharma contributed equally to this work and listed in alphabetical order.}
Video recognition has been advanced in recent years by benchmarks with rich annotations. However, research is still mainly limited to human action or sports recognition - focusing on a highly specific video understanding task and thus leaving a significant gap towards describing the overall content of a video. We fill  this gap by presenting a large-scale ``Holistic Video Understanding Dataset"~(HVU).
HVU is organized hierarchically in a semantic taxonomy that focuses on multi-label and multi-task video understanding as a comprehensive problem that encompasses the recognition of multiple semantic aspects in the dynamic scene. HVU contains approx.~572k videos in total  with 9 million annotations for training, validation and test set  spanning  over 3142  labels. HVU encompasses  semantic aspects defined on categories of scenes, objects, actions, events, attributes  and concepts which naturally captures the real-world scenarios.

We demonstrate the generalisation capability of HVU on three challenging tasks: 1.) Video classification,  2.) Video captioning and 3.) Video clustering tasks. In particular for video classification, we introduce a new spatio-temporal deep neural network architecture called ``Holistic Appearance and Temporal Network"~(HATNet) that builds on fusing 2D and 3D architectures into one by combining intermediate representations of appearance and temporal cues. HATNet focuses on the multi-label and multi-task learning problem and is trained in an end-to-end manner. 
Via our experiments, we validate the idea that holistic representation learning is complementary, and can play a key role in enabling many real-world applications. \textcolor{magenta}{\url{https://holistic-video-understanding.github.io/}}

\end{abstract}
\vspace{-0.2cm}

\input{intro_related.tex}

\input{HVU_Dataset/HVU_Dataset.tex}


\input{method.tex}
\input{experiment.tex}

\section{Conclusion}
This work presents the ``Holistic Video Understanding Dataset"~(HVU), a large-scale multi-task, multi-label video benchmark dataset with comprehensive tasks and annotations.  It contains 572k videos in total with 9M annotations, which is richly labeled over {3142} labels encompassing scenes, objects, actions, events, attributes  and concepts categories. Through our experiments, we show that the HVU can play a key role in learning a generic video representation via demonstration on three real-world tasks: video classification, video captioning and video clustering. Furthermore, we present a novel network architecture, HATNet, that combines 2D and 3D ConvNets in order to learn a robust spatio-temporal feature representation via multi-task and multi-label learning in an end-to-end manner. We believe that our work will inspire new research ideas for holistic video understanding. For the future plan, we are going to expand the dataset to $1$ million videos with similar rich semantic labels and also provide annotations for other important tasks like activity and object detection and video captioning.



\textbf{Acknowledgements:} This work was supported by DBOF PhD scholarship \& GC4 Flemish AI project, and the ERC Starting Grant ARCA (677650). We also would like to thank Sensifai for giving us access to the Video Tagging API for dataset preparation.

{\small
\bibliographystyle{splncs04}
\bibliography{egbib}
}

\end{document}

\pagestyle{headings}
\mainmatter
\def\ECCVSubNumber{3994}  

\title{Supplementary Material:\\ Large Scale Holistic Video Understanding} 

\titlerunning{Supplementary Material: Large Scale Holistic Video Understanding}
\authorrunning{Diba, Fayyaz, Sharma et al.}
\author{
    Ali Diba$^{1,5\star}$, Mohsen Fayyaz$^{2,\star}$, Vivek Sharma$^{3,\star}$,  \\ Manohar Paluri,  J\"{u}rgen Gall$^{2}$, Rainer Stiefelhagen$^{3}$, Luc Van Gool$^{1,4,5}$
}
\institute{$^{1}$KU Leuven, $^{2}$University of Bonn,$^{3}$KIT, Karlsruhe,  $^{4}$ETH Z\"{u}rich, $^{5}$Sensifai \\ 
\tt\small \{firstname.lastname\}@kuleuven.be, \{lastname\}@iai.uni-bonn.de, \{firstname.lastname\}@kit.edu, Balamanohar@gmail.com}

\maketitle


\appendix
\textbf{Appendix:}
This document provides supplementary material as mentioned in the main paper.
\section{HVU Dataset}
\footnote[]{$^{\star}$Ali Diba, Mohsen Fayyaz and Vivek Sharma contributed equally to this work and listed in alphabetical order.}
\subsection{Human Annotation Details}
The row machine generated annotations consist almost 8K labels. The initial stage of human verification on validation set resulted in 4378 labels. And the final stage of complete human verification/modification process ended up in 3142 labels. In human annotation process, 80 new labels are added by human annotators.

In specific for the HVU human verification task, we employed three different teams (Team-A, Team-B and Team-C) of 55 human annotators. Team-A works on the taxonomy of the dataset. This team builds the taxonomy based on the visual meaning and definition of the tags obtained from APIs prediction. Team-B and Team-C are the verification teams and perform four tasks. The tasks they performs are: (a) verify the tags of videos by watching each video and flag false tags; (b) review the tags by watching the videos of each tag and flag the wrong videos; (c) add tags to the videos if some tags are missing;
and (d) they suggest modification on tags such as, renaming or merging.

To make sure both Team-B and Team-C have a clear understanding of the tags and the corresponding videos, we ask them to use the provided tags definition from Team-A. For the aforementioned four tasks, Team-B goes through all the videos and provides the first round of clean annotations. Followed by this, Team-C reviews the annotations from Team-B to guarantee an accurate and cleaner version of annotations. Finally, Team-A reviews the suggestions provided from tasks (c) and (d) and apply them to the dataset.
The verification process takes $\sim$100 seconds on average per video clip for a trained worker.
It took about 8500 person-hours to firstly clean the machine-generated tags and remove errors and secondly add any possible missing labels from the dictionary.     
By incorporating the machine generated tags and human annotation, the HVU dataset covers a diverse set of tags with clean annotations. Using machine generated tags in the first step helps us to cover larger number of tags than a human can remember and label it in a reasonable time. 

To make sure that we have a balanced distribution of samples per tag, we consider a minimum number of 50 samples.

\begin{figure*}[t]
    \centering
    \hspace{0px}
    \includegraphics[scale=0.186]{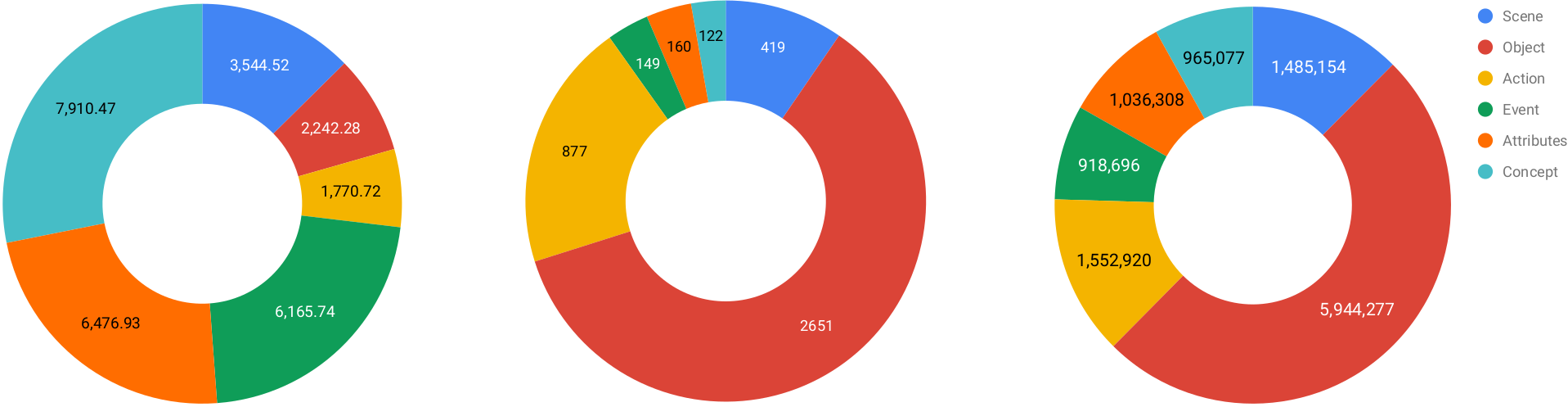}
    \vspace{-0.45cm}
    \caption{Left: Average number of samples per label in each of main categories. Middle: Number of labels for each main category. Right: Number of samples per main category. All statistics are for the machine generated tags of HVU training set.}
    \label{fig:ml_chart_cats}
\end{figure*}

\begin{figure*}[t]
    \centering
    \hspace{0px}
    \includegraphics[scale=0.213]{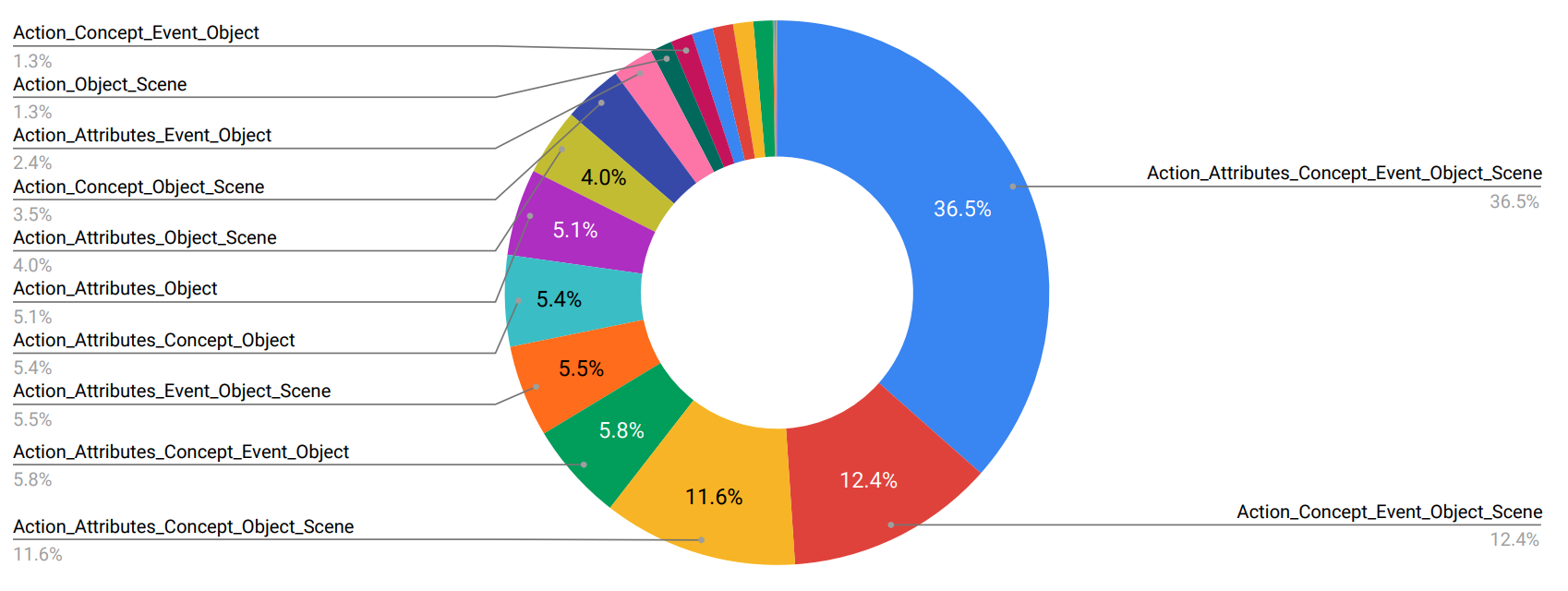}
    \vspace{-0.45cm}
    \caption{Coverage of different subsets of the 6 main semantic categories in videos. $16.4\%$ of the videos have annotations of all categories. All statistics are for the machine generated tags of HVU training set.}
    \label{fig:ml_chart_interscetion}
\end{figure*}

To provide more details regarding the HVU human annotation process, we report the statistics of the different stages of the annotation process. Table \ref{tab:HVU_ML_Cats} shows the statistics of the machine generated annotations of training set. Note, that the labels and categories are result of the initial human annotation process over the validation set of the dataset. 
The category with the highest number of labels and annotations is the object category. Concept is the category with the lowest number of labels. 
To have a better understanding of the statistics of the annotations we depict the distribution of categories with respect to the number of annotations, labels, and annotations per label in Figure \ref{fig:ml_chart_cats}. We can observe that the object category has the highest quota of labels and annotations, which is due to the abundance of objects in video. Despite having the highest quota of the labels and annotations, the object category does not have
the highest annotations per label ratio.
Figure \ref{fig:ml_chart_interscetion} shows the percentage of the different subsets of the main categories. There are 50 different sets of videos based on assigned semantic categories. About $36\%$ of the videos have all of the categories.

\begin{table*}[t] 
\centering
\small
\tabcolsep=0.2cm
\resizebox{12cm}{!}{
\begin{tabular}{  l|ccccccc } 
\toprule
Task Category & Scene & Object & Action & Event & Attribute & Concept & Total\\
 \midrule
 \midrule
 \#Labels & 419 & 2651 & 877 & 149 & 160 & 122 & 4378 \\
 \midrule
 \#Annotations & 1,485,154& 5,944,277 & 1,552,920 & 918,696 & 1,036,308 & 965,077 & 11,902,432\\
 \midrule
 \#Videos & 366,941& 480,821& 481,418& 320,428& 368,668& 375,664 & 481,418\\
 \bottomrule
\end{tabular}}
\caption{Statistics of machine generated tags of HVU training set for different categories. The category with the highest number of labels and annotations is the object category.}
\label{tab:HVU_ML_Cats}
\vspace{-0.6cm}
\end{table*}

\begin{figure}[t]
 \centering
 \resizebox{11cm}{!}{
 \includegraphics[width=1\columnwidth]{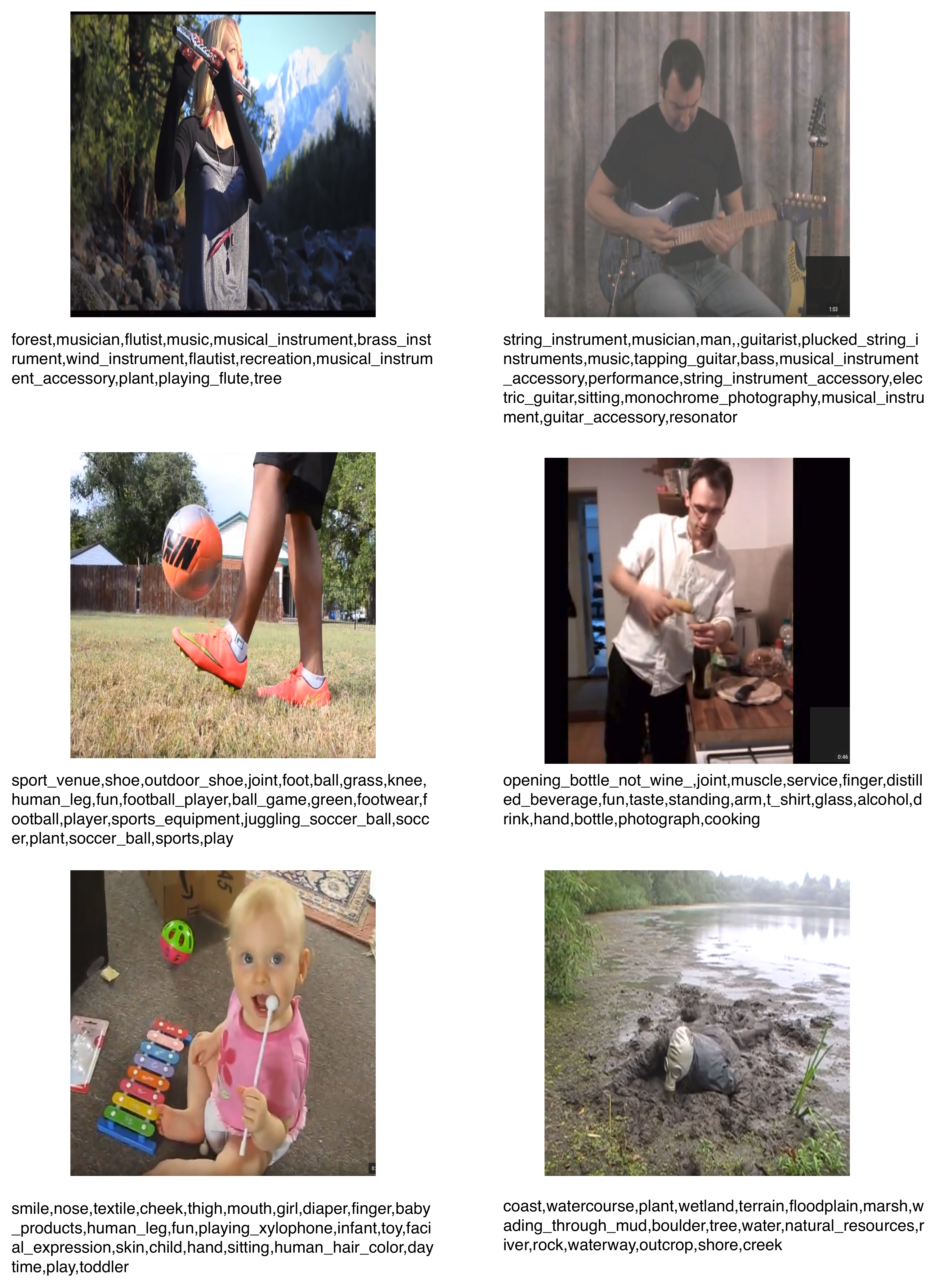}
 \vspace{-0.35cm}}
 \caption{Video frame samples from HVU with corresponding tags of different categories.} 
  \label{fig:fig1}
  \vspace{-0.7cm}
\end{figure}

\begin{figure}[t]
 \centering
 \resizebox{11cm}{!}{
 \includegraphics[width=1\columnwidth]{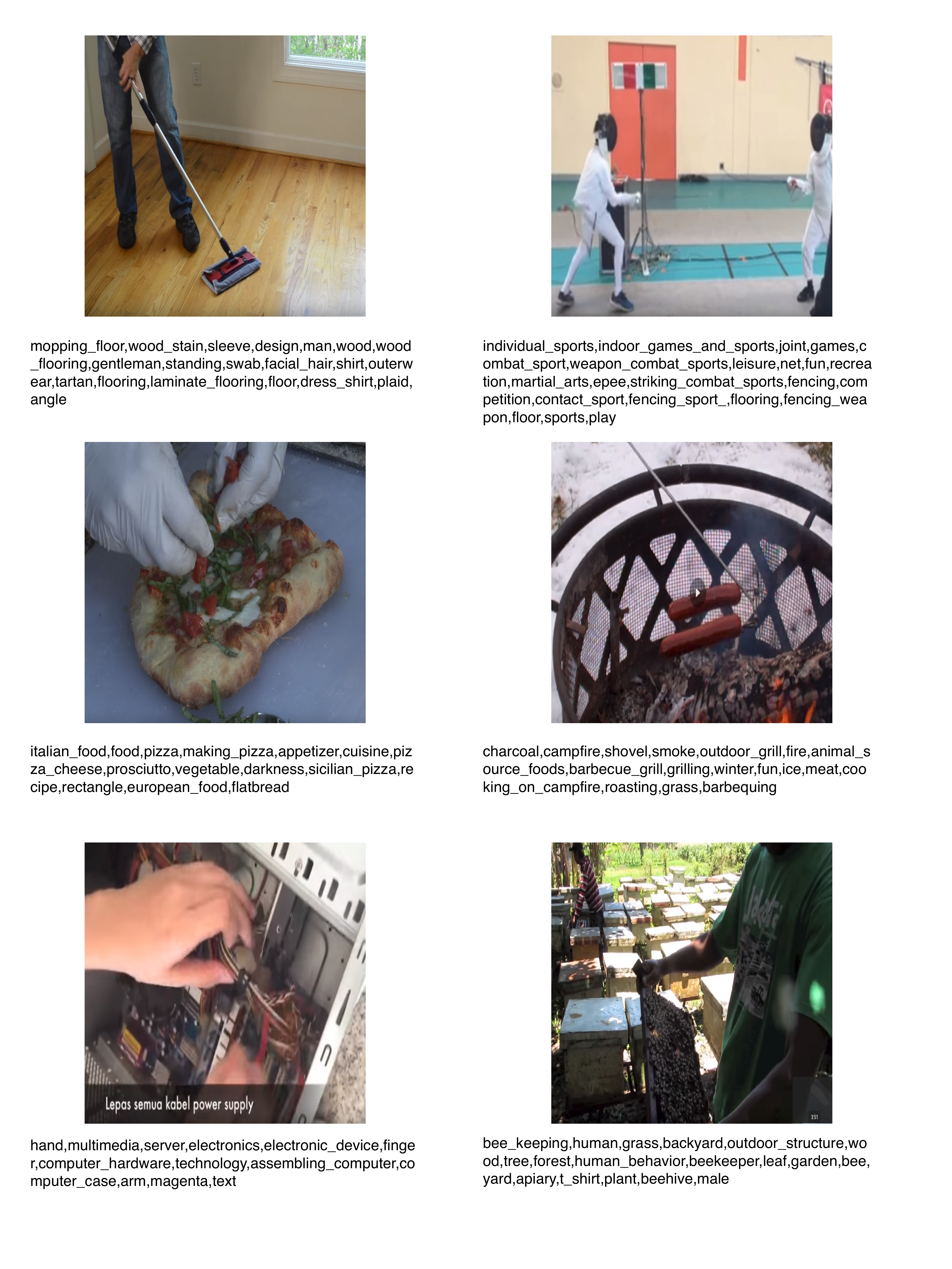}
 \vspace{-0.35cm}}
 \caption{More examples of video frame samples from HVU with corresponding tags of different categories.} 
  \label{fig:fig2}
\end{figure}

\subsection{Effect of Human Annotation}
To present the impact of human annotation process, we have evaluated both versions of the HVU with machine-generated tags and human-annotated tags. We have trained two 3D-ResNet18 for each set and the comparison came in Table~\ref{table:table1}.
\\

\begin{table*}[t] 
\centering
\small
\tabcolsep=0.13cm
\resizebox{12cm}{!}{
\begin{tabular}{  l|cccccc|c } 
\toprule
Dataset & Scene & Object & Action & Event & Attribute & Concept & HVU Overall $\%$\\
 \midrule
 \midrule
 Machine-Generated HVU & 46.3 &	22.4&	43.8&	31.4&	25.3&	20.1& 31.6 \\
 \midrule
 Human-Annotation  HVU & 50.1 &	27.9&	46.7&	35.7&	29.2&	23.2& 35.4 \\
 \bottomrule
\end{tabular}}
\caption{Performance comparison between machine generated and human-verified tags of HVU. This evaluation shows how human annotation process is crucial to have a more efficient dataset. The CNN model which is used for this experiment is 3D-ResNet18.}
\label{table:table1}
\end{table*}


\subsection{HVU Samples}
We present some samples of videos and their corresponding tags in Fig~\ref{fig:fig1} and Fig~\ref{fig:fig2}.

\subsection{Effect of Additional Categories on Kinetics}
One of our arguments in our paper is about how more semantic categories like object, scene, etc can lead to learn effective video representation. We have shown results on the HVU dataset in the paper. Here, we provided the similar experiment for the Kinetics-600 as a subset of our HVU.
We have compared performance of a 3D-ResNet18 trained on Kinetics videos with its action labels versus trained on full HVU labels for the same videos. For the evaluation, we have measured the performance on Kinetics action labels. It can be seen in Table~\ref{table:kin} that having more semantic labels in the training for Kinetics, improves the action classification performance. It is due to the fact that HVU can bring more capabilities to the deep models for learning new visual features for understanding  videos.

\begin{table*}[t] 
\centering
\small
\tabcolsep=0.15cm
\resizebox{8cm}{!}{
\begin{tabular}{ l|c  }
\toprule
Training Labels   & Action Recognition Performance \\
\hline 
\hline 
Action  & 65.6\\
Action + HVU  & 68.8  \\
\bottomrule
\end{tabular}}
\caption{Evaluation of training Kinetics with HVU labels.}
\label{table:kin}
\end{table*}






%% file: intro_related.tex


\begin{figure}[t]
 \centering
 \resizebox{11cm}{!}{
 \includegraphics[width=1\columnwidth]{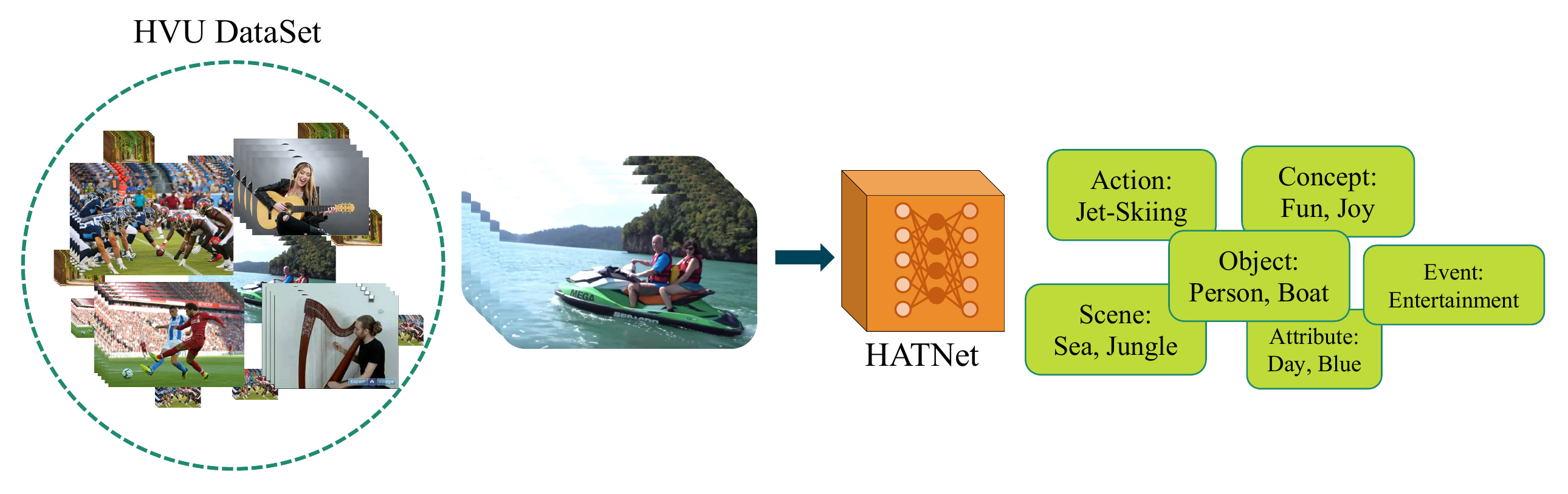}
 \vspace{-0.35cm}}
 \caption{{Holistic Video Understanding Dataset: A multi-label and multi-task fully annotated dataset and HATNet as a new deep ConvNet for video classification.}} 
  \label{fig:fig1}
  \vspace{-0.7cm}
\end{figure}

\section{Introduction}

Video understanding is a comprehensive problem that encompasses the recognition of multiple semantic aspects that include: a scene or an environment, objects, actions, events, attributes, and concepts. Even if considerable progress is made in video recognition, it is still rather limited to action recognition - this is due to the fact that there is no established video benchmark that integrates joint recognition of multiple semantic aspects in the dynamic scene. While Convolutional Networks(ConvNets) have caused several sub-fields of computer vision to leap forward, one of the expected drawbacks of training the ConvNets for video understanding  with a single label per task is insufficiency to describe  the content of a video.
This issue primarily impedes the ConvNets to learn a generic feature representation towards challenging holistic video analysis. To this end, one can easily overcome this issue by recasting the video understanding problem as multi-task classification, where multiple labels are assigned to a video from multiple semantic aspects.
Furthermore, it is possible to learn a generic feature representation for video analysis and understanding. This is in line with 
image classification ConvNets trained on ImageNet that facilitated the learning of generic feature representation for several vision tasks. Thus, training ConvNets on a multiple semantic aspects dataset can be directly applied for  holistic recognition and understanding of concepts in video data, which makes it very useful to describe  the content of a video.

To address the above drawbacks, this work presents the ``Holistic Video Understanding Dataset"~(\textbf{HVU}). HVU is organized hierarchically in a semantic taxonomy that aims at providing a multi-label and multi-task large-scale video benchmark with a comprehensive list of tasks and annotations for video analysis and understanding. HVU dataset consists of 476k, 31k and 65k samples in train, validation and test set, and is a sufficiently large dataset, which means that the scale of  dataset approaches that of image datasets.
HVU  contains approx. 572k videos in total, with $\sim$7.5M annotations for training set, $\sim$600K for validation set, and $\sim$1.3M for test set spanning over 3142 labels. A full spectrum encompasses the recognition of multiple semantic aspects defined on them including  248 categories for scenes, 1678 for objects, 739 for actions, 69 for events, 117 for attributes and 291 for concepts, which naturally captures the long tail distribution of visual concepts in the real world problems. All these tasks are supported by rich annotations with an average of 2112 annotations per label. The HVU action categories builds on action recognition datasets~\cite{ava,kinetics,hmdb51,ucf101,hacs} and further extend them by incorporating labels of scene,  objects, events, attributes, and concepts in a video. The above thorough annotations enable developments of strong algorithms for a holistic video understanding to describe the content of a video. Table~\ref{tab:HVU_Cats} shows the dataset statistics.

In order to show the importance of holistic representation learning, we demonstrate the influence of HVU on three challenging tasks:  video classification, video captioning and video clustering. Motivated by holistic representation learning, for the task of video classification, we introduce a new spatio-temporal architecture called ``Holistic Appearance and Temporal Network"~(HATNet) that focuses on the multi-label and multi-task 
learning for jointly solving multiple spatio-temporal problems simultaneously. HATNet fuses 2D and 3D architectures into one by combining intermediate representations of appearance and temporal cues, leading to a robust spatio-temporal representation. Our HATNet is evaluated on challenging video classification datasets, namely HMDB51, UCF101 and Kinetics. We experimentally show that our HATNet achieves outstanding results. Furthermore, we show the positive effect of training models using more semantic concepts on transfer learning. In particular, we show that pre-training the model on HVU with more semantic concepts improves the fine-tuning results on other datasets and tasks compared to pre-training on single semantic category datasets such as, Kinetics.
This shows the richness of our dataset as well as the importance of multi-task learning. Furthermore, our experiments on video captioning and video clustering demonstrates the generalisation capability of HVU on other tasks by showing promising results in comparison to the state-of-the-art.

\section{Related Work}

\begin{figure*}[t]
    \centering
     \includegraphics[scale=0.236]{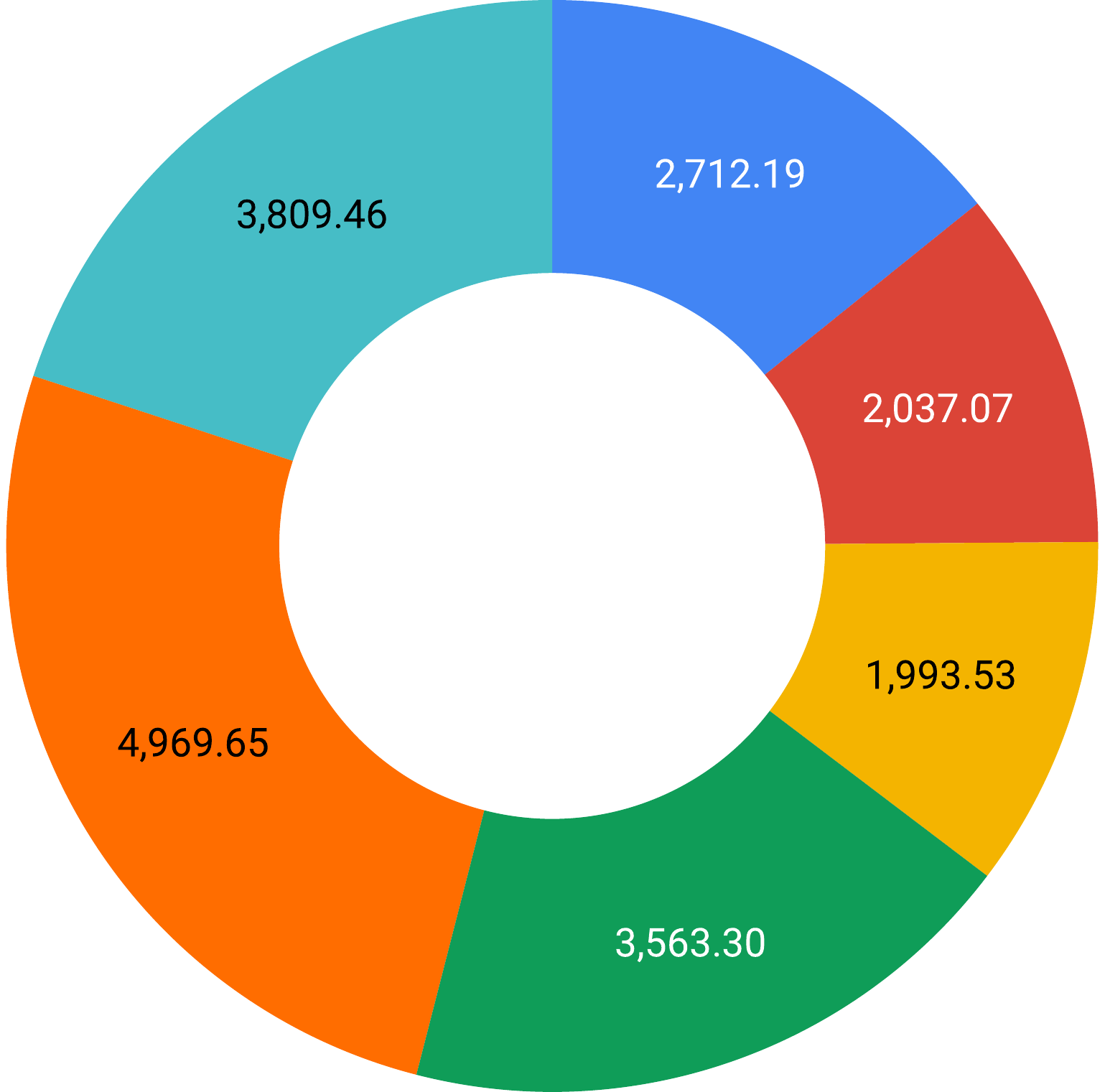}
     \hspace{0.05cm}
     \includegraphics[scale=0.236]{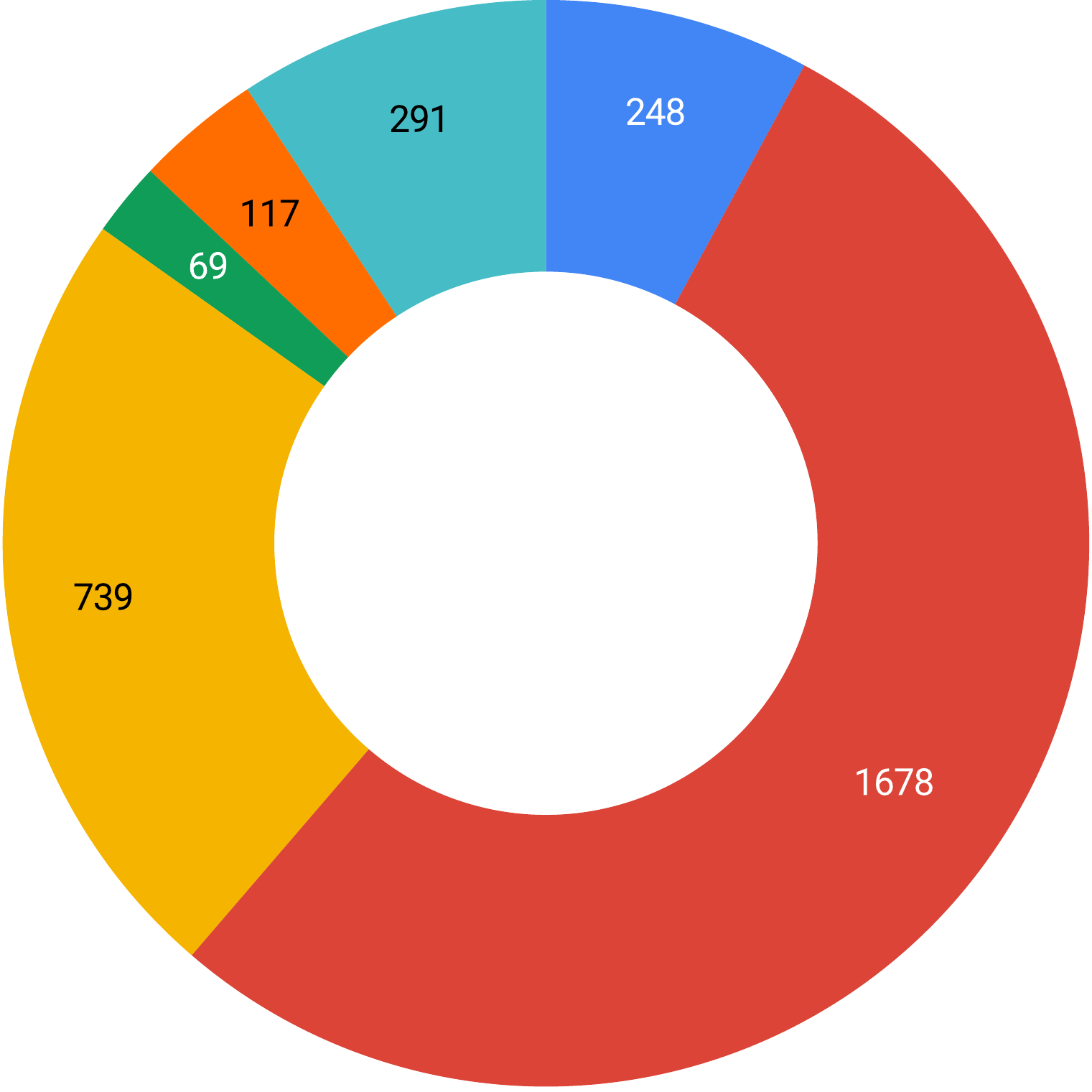} 
     \includegraphics[scale=0.236]{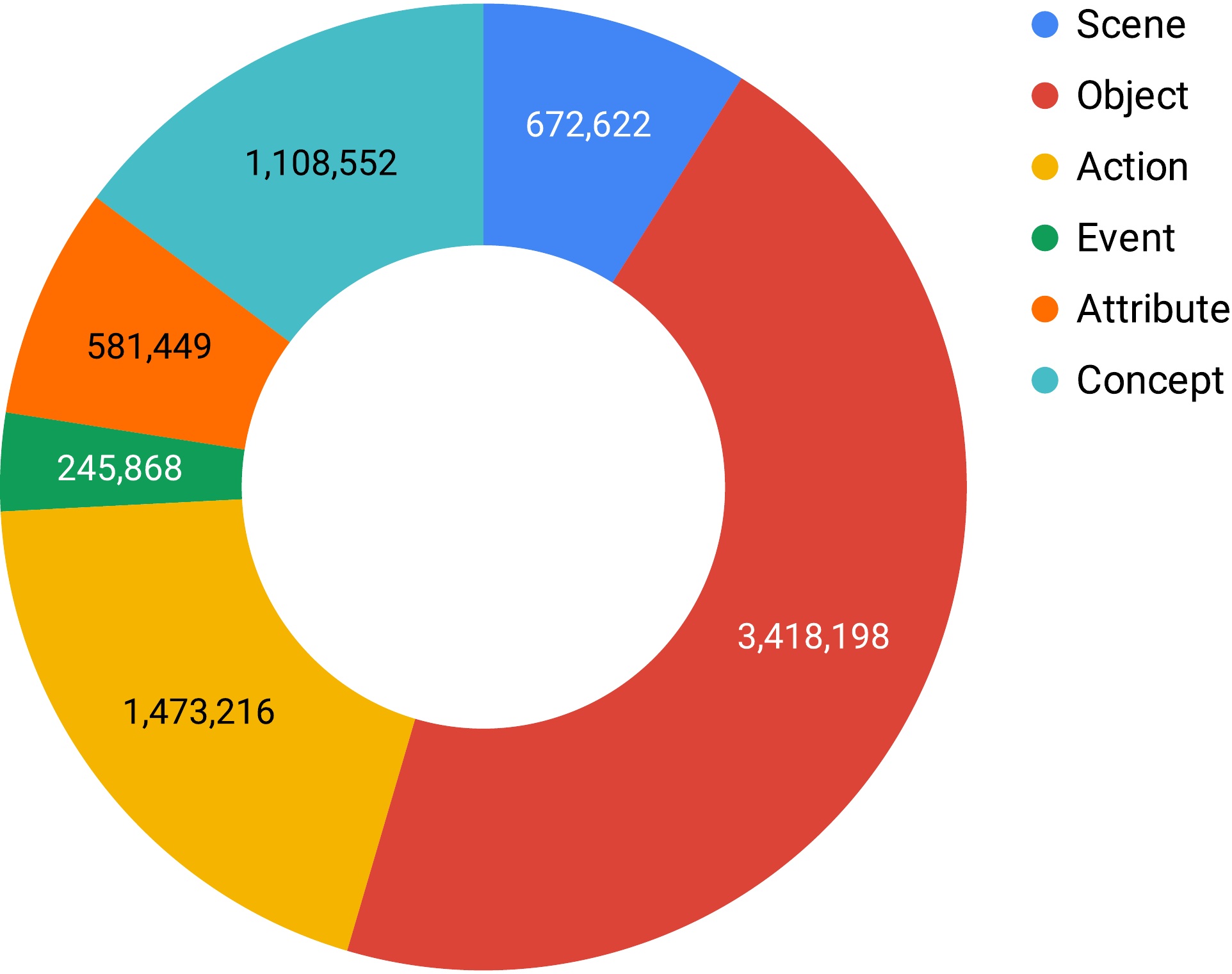}
     \vspace{-0.35cm}
    \caption{Left: Average number of samples per label in each of main categories. Middle: Number of labels for each main category. Right: Number of samples per main category.}
    \label{fig:charts_clc_ant}
    \vspace{-0.5cm}
\end{figure*}

\begin{figure*}[t]
    \centering
    \includegraphics[scale=0.304]{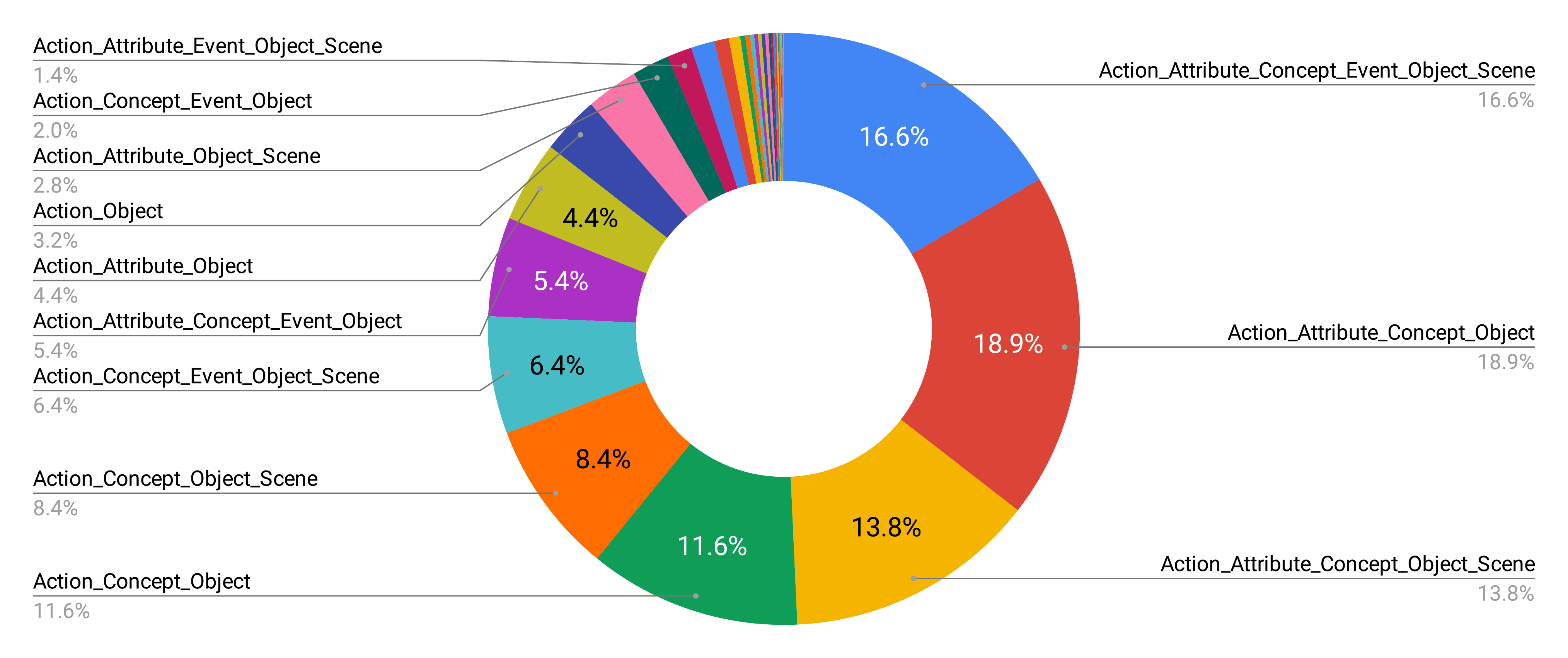}
    \vspace{-0.45cm}
    \caption{Coverage of different subsets of the 6 main semantic categories in videos. $16.6\%$ of the videos have annotations of all categories.}
    \label{fig:chart_interscetion}
    \vspace{-0.7cm}
\end{figure*}

\noindent
\textbf{Video Recognition with ConvNets:} 
As to prior hand-engineered~\cite{mbh,hog3d,hof,sift3d,idt,esurf} and low-level temporal structure~\cite{basura,actoms,temporalactiondecomposition,dynamicposelets} descriptor learning there
is a vast literature and is beyond the scope of this paper.

Recently ConvNets-based action recognition~\cite{pooling,sport1m,twostream,c3d,tsn} has taken a leap to exploit the appearance and the temporal information. These methods operate on 2D (individual image-level)~\cite{tle,lstm1,actionvlad,sun3d,fishernet,tsn,n3d} or 3D (video-clips or snippets of $K$ frames)~\cite{pooling,c3d,res3d,R2+1}. The filters and pooling kernels for these architectures are 3D (x, y, time) i.e. 3D convolutions~($s \times s\times d$)~\cite{n3d} where $d$ is the kernel's temporal depth and $s$ is the kernel's spatial size. These 3D ConvNets are intuitively effective because such 3D convolution can be used to directly extract spatio-temporal features from raw videos. Carreira\textit{et al.} proposed inception~\cite{inception} based 3D CNNs, which they referred to as I3D~\cite{i3d}. More recently, some works introduced temporal transition layer that models variable temporal convolution kernel depths over shorter and longer temporal ranges, namely T3D~\cite{t3d}. Further Diba\textit{et al.}~\cite{stcnn} propose spatio-temporal channel correlation that models correlations between channels of a 3D ConvNets wrt. both spatial and temporal dimensions. In contrast to these prior works, our work differs substantially in scope and technical approach. We propose an architecture, HATNet, that exploits both 2D ConvNets and 3D ConvNets to learn an effective  spatio-temporal feature representation. Finally, it is worth noting the self-supervised ConvNet training works from unlabeled sources~\cite{disinit,sharmatbiom,sharmavo}, such as Fernando\textit{et al.}~\cite{Odd-one} and Mishra\textit{et al.}~\cite{Shuffle} generate training data by shuffling the video frames; Sharma\textit{et al.}~\cite{sharmaacmmm,sharmacvpr17,sharmafg,sharmafg20} mines labels using a distance matrix or clustering based on similarity although for video face clustering; Wei\textit{et al.}~\cite{wei} predict the ordering task; Ng\textit{et al.}~\cite{actionflow} estimates optical flow while recognizing actions; Diba\textit{et al.}~\cite{dynamonet} predicts short term future frames while recognizing actions. Self-supervised and unsupervised representation learning is beyond the scope of this paper. 

The closest work to ours is by Ray\textit{et al.}~\cite{soadata}. Ray\textit{et al.} concatenate pre-trained deep features, learned independently for the different tasks, scenes, object and actions aiming to the recognition, in contrast our HATNet is trained end-to-end for multi-task and multi-label recognition in videos.

\noindent
\textbf{Video Classification Datasets:}  
Over the last decade, several video classification datasets~\cite{mpii,actnet,hmdb51,kth,ucf101} have been made publicly available with a focus on action recognition, as summarized in Table~\ref{table:Action_Datasets}. We briefly review some of the most influential action datasets available.  The HMDB51~\cite{hmdb51} and UCF101~\cite{ucf101} has been very important in the field of action recognition. However, they are simply not large enough for training deep ConvNets from scratch. Recently, some large action recognition datasets were introduced, such as ActivityNet~\cite{actnet} and Kinetics~\cite{kinetics}. ActivityNet contains 849 hours of videos, including 28,000 action instances.  Kinetics-600 contains 500k videos spanning 600 human action classes with more than 400 examples for each class. The current experimental strategy is to first pre-train models on these large-scale video datasets~\cite{actnet,sport1m,kinetics} from scratch and then fine-tune them on small-scale datasets~\cite{hmdb51,ucf101} to analyze their transfer behavior. Recently, a few other action datasets have been introduced  with more samples, temporal duration  and the diversity of category taxonomy, they are HACS~\cite{hacs}, AVA~\cite{ava}, Charades~\cite{Charades} and Something-Something~\cite{smt-smt}. Sports-1M~\cite{sport1m} and YouTube-8M~\cite{youtube8m} are the video datasets with million-scale samples. They consist quite longer videos rather than the other datasets and their annotations are provided in video-level and not temporally stamped. YouTube-8M labels are machine-generated without any human verification in the loop and Sports-1M is just focused on sport activities.

\begin{table*}[t] 
\centering
\small
\tabcolsep=0.2cm
\resizebox{12cm}{!}{
\begin{tabular}{  l|ccccccc } 
\toprule
Task Category & Scene & Object & Action & Event & Attribute & Concept & Total\\
 \midrule
 \midrule
 \#Labels & 248 & 1678 & 739 & 69 & 117 & 291 & 3142 \\
 \midrule
 \#Annotations & 672,622& 3,418,198 & 1,473,216 & 245,868 & 581,449 & 1,108,552 & 7,499,905\\
 \midrule
 \#Videos & 251,794& 471,068& 479,568& 164,924& 316,040& 410,711& 481,417\\
 \bottomrule
\end{tabular}}
\caption{Statistics of the HVU training set for different categories. The category with the highest number of labels and annotations is the object category.}
\label{tab:HVU_Cats}
\vspace{-0.6cm}
\end{table*}

\begin{table*}[t] 
\centering
\small
\tabcolsep=0.2cm
\resizebox{12cm}{!}{
\begin{tabular}{  l|ccccccc c } 
\toprule
Dataset & Scene & Object & Action & Event & Attribute & Concept &  \#Videos & Year\\
\hline 
\hline 
HMDB51~\cite{hmdb51} & - & - & 51 & - & - & - & ~7K& '11\\
\hline 
UCF101~\cite{ucf101} & - & - & 101 & - & - & - & ~13K& '12\\
\hline 
ActivityNet~\cite{actnet} & - & - & 200 & - & - & - & ~20K& '15\\
\hline 
AVA~\cite{ava} & - & - & 80 & - & - & - & 57.6K& '18\\
\hline 
Something-Something~\cite{smt-smt} & - & - & 174 & - & - & - & 108K& '17\\
\hline 
HACS~\cite{hacs} & - & - & 200 & - & - & - & 140K& '19\\
\hline 
Kinetics~\cite{kinetics} & - & - & 600 & - & - & - & ~500K& '17\\
\hline 
EPIC-KITCHEN~\cite{epickitchen} & - & 323 & 149& - & - & - & 39.6K& '18\\
\hline
SOA~\cite{soadata} & 49 &356 & 148& - & - & - & 562K& '18\\
\hline 
\hline 
HVU~(\textbf{Ours})& 248 & 1678 & 739 & 69 & 117 & 291 & 572K& '20\\
\bottomrule
\end{tabular}}
\caption{Comparison of the HVU dataset with other publicly available video recognition datasets in terms of \#labels per category. Note that SOA is not publicly available.}
\label{table:Action_Datasets}
\vspace{-1cm}
\end{table*}

A similar spirit of HVU is observed in SOA dataset~\cite{soadata}. SOA aims to recognize visual concepts, such as scenes, objects and actions. In contrast, HVU has several orders of magnitude more semantic labels(6 times larger than SOA) and not just limited to scenes, objects, actions only, but also including events, attributes, and concepts. Our HVU dataset can help the computer vision community and bring more attention to holistic video understanding  as a comprehensive, multi-faceted problem. Noticeably, the SOA paper was published in 2018, however the dataset is not released while our dataset is ready to become publicly available.


Motivated by efforts in large-scale benchmarks for object recognition in static images, i.e. the Large Scale Visual Recognition Challenge~(ILSVRC) to learn a generic feature representation is now a back-bone to support several related vision tasks. We are driven by the same spirit towards learning a generic feature representation at the video level for holistic video understanding.
\vspace{-0.2cm}

%% file: HVU_Dataset/HVU_Dataset.tex
\section{HVU Dataset}

The HVU dataset is organized hierarchically in a semantic taxonomy of holistic video understanding. Almost all real-wold conditioned video datasets are targeting human action recognition. However, a video is not only about an action which provides a human-centric description of the video. By focusing on human-centric descriptions, we ignore the information about scene, objects, events and also attributes of the scenes or objects available in the video. 
While SOA~\cite{soadata} 
has categories of scenes, objects, and actions, to our knowledge it is not publicly available. 
Furthermore, HVU 
has more categories as it is shown in Table~\ref{table:Action_Datasets}. 
 One of the important research questions which is not addressed well in recent works on action recognition, is leveraging the other contextual information in a video. 
The HVU dataset makes it possible to assess the effect of learning and knowledge transfer among different tasks, such as enabling transfer learning of  object recognition in videos to action recognition and vice-versa.
In summary, HVU can help the vision community and bring more interesting solutions to holistic video understanding. Our dataset focuses on the recognition of scenes, objects, actions, attributes, events, and concepts in user generated videos. Scene, object, action and event categories definition is the same and standard as in other image and datasets. For attribute labels, we target attributes describing scenes, actions, objects or events. The concept category refers to any noun and label which present a grouping definition or related higher level in the taxonomy tree for labels of other categories.

\subsection{HVU Statistics}
HVU consists of \textbf{572k} videos. The number of video-clips for train, validation, and test set are \textbf{481k}, \textbf{31k} and \textbf{65k} respectively.
The dataset consists of trimmed video clips. 
In practice, the duration of the videos are different with a maximum of $10$ seconds length. HVU has $6$ main categories: scene, object, action, event, attribute, and concept. In total, there are {3142} labels with approx. 7.5M annotations for the training, validation and test set. On average, there are $\sim$2112 annotations per label. We depict the distribution of categories with respect to the number of annotations, labels, and annotations per label in Fig.~\ref{fig:charts_clc_ant}. We can observe that the object category has the highest quota of labels and annotations, which is due to the abundance of objects in video. Despite having the highest quota of the labels and annotations, the object category does not have the highest annotations per label ratio. However, the average number of $\sim$2112 annotations per label is a reasonable amount of training data for each label. The scene category does not have a large amount of labels and annotations which is due to two reasons: the trimmed videos of the dataset and the short duration of the videos. This distribution is somewhat the same for the action category. The dataset statistics for each category are shown in Table~\ref{tab:HVU_Cats} for the training set.

\subsection{Collection and Annotation}
\label{subsec:collection_and_annotation}
Building a large-scale video understanding dataset is a time-consuming task. In practice, there are two main tasks which are usually most time consuming for creating a large-scale video dataset: (a) data collection and (b) data annotation.
Recent popular datasets, such as ActivityNet, Kinetics, and YouTube-8M are collected from Internet sources like YouTube.
For the annotation of these datasets, usually a semi-automatic crowdsourcing strategy is used, in which a human manually verifies the crawled videos from the web. We adopt a similar strategy with difference in the technical approach
to reduce the cost of data collection and annotation. Since, we are 
interested in the user generated videos, thanks to the taxonomy diversity of YouTube-8M~\cite{youtube8m}, Kinetics-600~\cite{kinetics} and HACS~\cite{hacs}, we use these datasets as main source of the HVU. 
By using these datasets as the source, we also do not have to deal with copyright or privacy issues so we can publicly release the dataset. Moreover, this ensures that none of the test videos of existing datasets is part of the training set of HVU. Note that, all of the aforementioned datasets are action recognition datasets.

Manually annotating a large number of videos with multiple semantic categories (i.e thousands of concepts and tags) has two major shortcomings, (a) manual annotations are error-prone because a human cannot be attentive to every detail occurring in the video that leads to mislabeling and are difficult to eradicate; (b) large scale video annotation in specific is a very time consuming task due to the amount and temporal duration of the videos. To overcome these issues, we employ a two-stage framework for the HVU annotation. In the first stage, we utilize the Google Vision API~\cite{googleAI} and Sensifai Video Tagging API~\cite{sensifai} to get rough annotations of the videos. The APIs predict 30 tags per video. We keep the probability threshold of the APIs relative low ($\sim$~30\%) as a guarantee to avoid false rejects of tags in the video.  The tags were chosen from a dictionary with almost 8K words. This process resulted in almost 18 million tags for the whole dataset. In the second stage, we apply human verification to remove any possible mislabeled noisy tags and also add possible missing tags missed by the APIs from some recommended tags of similar videos. The human annotation step resulted in 9 million tags for the whole dataset with $\sim$3500 different tags.

We provide more detailed statistics and discusion regarding the annotation process in the supplementary materials.

\subsection{Taxonomy}
Based on the predicted tags from the Google and the Sensifai APIs, we found that the number of obtained tags is approximately $\sim$8K before cleaning. The services can recognize videos with tags spanning over categories of scenes, objects, events, attributes, concepts, logos, emotions, and actions. As mentioned earlier, we remove tags with imbalanced distribution and finally, refine the tags to get the final taxonomy by using the WordNet~\cite{wordnet} ontology. The refinement and pruning process aims to preserve the true distribution of labels. Finally, we ask the human annotators to classify the tags into 6 main semantic categories, which are scenes, objects, actions, events, attributes and concepts.

In fact, each video can be assigned to multiple semantic categories. 
Almost 100K of the videos have all of the semantic categories. In comparison to SOA, almost half of HVU videos have labels for scene, object and action together. Figure~\ref{fig:chart_interscetion} shows the percentage of the different subsets of the main categories.

 \begin{figure*}[t]
 \centering
 \includegraphics[width=0.99\columnwidth]{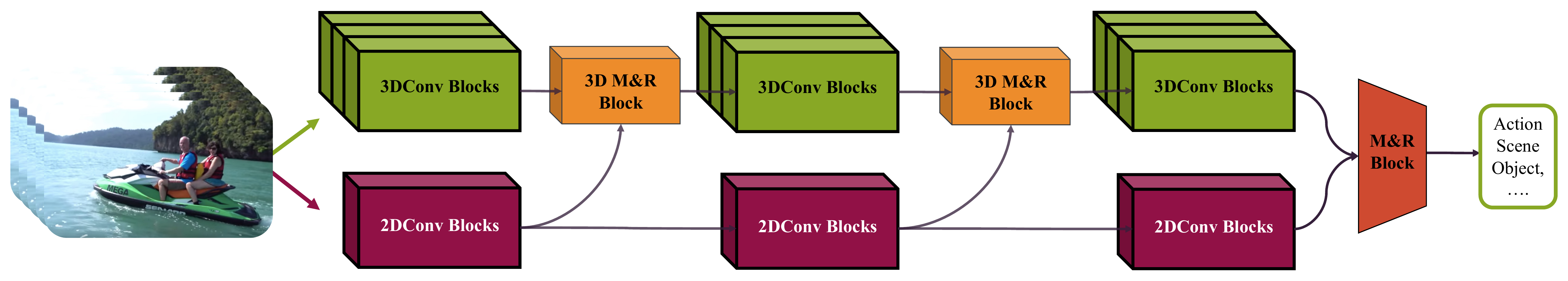}
 \vspace{-0.35cm}
 \caption{HATNet: A new 2D/3D deep neural network with 2DConv, 3DConv blocks and merge and reduction (M\&R) block to fuse 2D and 3D feature maps in intermediate stages of the network. HATNet combines the appearance and temporal cues with the overall goal to compress them into a more compact representation.} 
  \label{fig:HATNet}
  \vspace{-0.7cm}
 \end{figure*}

%% file: method.tex
\section{Holistic Appearance and Temporal Network}
We first briefly discuss state-of-the-art 3D ConvNets for video classification and then propose our new proposed ``Holistic Appearance and Temporal Network"~(HATNet) for multi-task and multi-label video classification.

\subsection{3D-ConvNets Baselines}
3D ConvNets are designed to handle temporal cues available in video clips and are shown to be efficient 
performance-wise for video classification. 3D ConvNets exploit both spatial and temporal information in one pipeline. In this work, we chose 3D-ResNet~\cite{res3d} and STCnet~\cite{stcnn} as our 3D CNNs baseline which have competitive results on Kinetics and UCF101. To measure the performance on the multi-label HVU dataset, we use mean average precision (mAP) over all labels. We also report the individual performance on each category 
separately. The comparison between all of the methods can be found in {Table~\ref{table:baselines}}. These networks are trained with binary cross entropy loss.

\subsection{Multi-Task Learning 3D-ConvNets} \label{subsec:multi_task_learning}
Another approach which is studied in this work to tackle the HVU dataset is to have the problem solved with multi-task learning or a joint training method. As we know the HVU dataset consists of high-level categories like objects, scenes, events, attributes, and concepts, so each of these categories can be dealt like separate tasks.
In our experiments, we have defined six tasks, scene, object, action, event, attribute, and concept classification. So our multi-task learning network is trained with six objective functions, that is with multi-label classification for each task. The trained network is a 3D-ConvNet which has separate Conv layers as separate heads for each of the tasks at the end of the network.

For each head we use the binary cross entropy loss since it is a multi-label classification for each of the categories.

\subsection{2D/3D HATNet}
Our ``Holistic Appearance and Temporal Network"~(HATNet) is a spatio-temporal neural network, which extracts temporal and appearance information in a novel way to maximize engagement of the two sources of information and also the efficiency of video recognition. The motivation of proposing this method is deeply rooted in a need of handling different levels of concepts in holistic video recognition. Since we are dealing with still objects, dynamic scenes, different attributes and also different human activities, we need a deep neural network that is able to focus on different levels of semantic information. We propose a flexible method to use a 2D pre-trained model on a large image dataset like ImageNet and a 3D pre-trained model on video datasets like Kinetics to fasten the process of training but the model can be trained from scratch as it is shown in our experiments as well. The proposed HATNet is capable of learning a hierarchy of spatio-temporal feature representation using appearance and temporal neural modules.

\textbf{Appearance Neural Module.} In HATNet design, 
we use 2D ConvNets with  2D Convolutional~(2DConv)  blocks to extract static cues of individual frames in a video-clip. Since we aim to recognize objects, scenes and attributes alongside of actions, it is necessary to have this module in the network which can handle these concepts better. Specifically, we use 2DConv to capture the spatial structure in the frame.

\textbf{Temporal Neural Module.} In HATNet architecture, the 3D Convolutions~(3DConv) module handles temporal cues dealing with interaction in a batch of frames. 3DConv aims to capture the relative temporal information between frames. It is crucial to have 3D convolutions in the network to learn relational motion cues for efficiently understanding dynamic scenes and human activities. We use ResNet18/50 for both the 3D and 2D modules,
so that they have the same spatial kernel sizes, and thus we can combine the output of the appearance and temporal branches at any intermediate stage of the network. 

Figure~\ref{fig:HATNet} shows how we combine the 2DConv and 3DConv branches and  use merge and reduction blocks to fuse feature maps at the intermediate stages of HATNet.  Intuitively, combining the appearance and temporal features are complementary for video understanding and this fusion step aims to compress them into a more compact and robust representation.
In the experiment section, we discuss in more detail about the HATNet design and how we apply merge and reduction modules between 2D and 3D neural modules.
Supported by our extensive experiments,
we show that HATNet complements the holistic video recognition, including understanding the dynamic and static aspects of a scene and also human action recognition. In our experiments, we have also performed tests on HATNet based multi-task learning similar to 3D-ConvNets based multi-task learning discussed in Section~\ref{subsec:multi_task_learning}. HATNet has some similarity to the SlowFast~\cite{slowfast} network but there are major differences. SlowFast uses two 3D-CNN networks for a slow and a fast branch. HATNet has one 3D-CNN branch to handle motion and dynamic information and one 2D-CNN to handle static information and appearance. HATNet also has skip connections with M\&R blocks between 3D and 2D convolutional blocks to exploit more information.

\textbf{2D/3D~HATNet Design.}
 The HATNet includes two branches: first is the 3D-Conv blocks with merging and reduction block and second branch is 2D-Conv blocks. 
After each 2D/3D blocks, we merge the feature maps from each block and perform a channel reduction by applying a $1 \times 1 \times 1$ convolution. Given the feature maps of the first block of both 2DConv and 3DConv, that have 64 channels each. We first concatenate these maps, resulting in 128 channels, and then apply $1 \times1 \times 1$ convolution with 64 kernels for channel reduction, resulting in an output with 64 channels.
The merging and reduction is done in the 3D and 2D branches, and continues independently until the last merging with two branches. 

We employ 3D-ResNet and STCnet~\cite{stcnn} with ResNet18/50 as the HATNet backbone in our experiments. The STCnet is a model of 3D networks with spatio-temporal channel correlation modules which improves 3D networks performance significantly. We also had to make a small change to the 2D branch and remove pooling layers right after the first 2D Conv to maintain a similar feature map size between the 2D and 3D branches since we use 112$\times$112 as input-size.

\begin{table*}[t] 
\centering
\small
\tabcolsep=0.15cm
\resizebox{12cm}{!}{
\begin{tabular}{  l|cccccc|c } 
\toprule
Model & Scene & Object & Action & Event & Attribute & Concept & HVU Overall $\%$\\
 \midrule
 \midrule
 3D-ResNet & 50.6 &	28.6&	48.2&	35.9&	29&	22.5& 35.8 \\
 \midrule
 3D-STCNet & 51.9&	30.1&	50.3&	35.8&	29.9&	22.7 & 36.7 \\
 \midrule
 HATNet    & \textbf{55.8}&	\textbf{34.2}&	\textbf{51.8}	&\textbf{38.5}&	\textbf{33.6}&	\textbf{26.1}& \textbf{40}\\
 \bottomrule
\end{tabular}}
\caption{MAP~(\%) performance of different architecture on the HVU dataset. The backbone ConvNet for all models is ResNet18.}
\label{table:baselines}
\end{table*}

%% file: experiment.tex

\begin{table}[t] 
\centering
\small
\tabcolsep=0.10cm
\begin{tabular}{ l|c c c c c c| l }
\toprule
Model &  Scene & Object& Action & Event & Attribute & Concept & Overall\\
\hline 
\hline 
3D-ResNet (Standard)  & 50.6 &	28.6&	48.2&	35.9&	29&	22.5& 35.8 \\
HATNet (Standard)& {55.8}&	{34.2}&	{51.8}	&{38.5}&{33.6}&{26.1}&{40}\\
\hline 
3D-ResNet (Multi-Task) & 51.7 & 29.6&  48.9& 36.6& 31.1 & 24.1& 37\\
HATNet (Multi-Task)& \textbf{57.2}& \textbf{35.1} & \textbf{53.5} & \textbf{39.8}& \textbf{34.9}& \textbf{27.3} & \textbf{41.3} \\
\bottomrule
\end{tabular}
\caption{Multi-task learning performance (mAP~(\%) comparison of 3D-ResNet18 and HATNet, when trained on HVU with all categories in the multi-task pipeline. The backbone ConvNet for all models is ResNet18.}
\label{table:multitask}
\vspace{-0.7cm}
\end{table}

\begin{table}[t] 
\centering
\small
\tabcolsep=0.15cm
\resizebox{8cm}{!}{
\begin{tabular}{ l|c c c }
\toprule
Pre-Training Dataset  & UCF101 & HMDB51 & Kinetics \\
\hline 
\hline 
From Scratch & 65.2 & 33.4 & 65.6\\
Kinetics & 89.8  & 62.1 & -\\
HVU & \textbf{90.5} & \textbf{65.1} & \textbf{67.8}\\
\bottomrule
\end{tabular}}
\caption{Performance (mAP~(\%)) comparison of HVU and Kinetics datasets for transfer learning generalization ability when evaluated on different action recognition dataset. The trained model for all of the datasets is 3D-ResNet18.}
\label{table:TFlearn}
\end{table}

\begin{table*}[t] 
\centering
\small
\tabcolsep=0.1cm
\resizebox{12cm}{!}{
\begin{tabular}{ l | c |  c | c  c  c  c}
\toprule
\textbf{Method} &  \textbf{Pre-Trained Dataset}& \textbf{CNN Backbone} &\textbf{UCF101} & \textbf{HMDB51} & \textbf{Kinetics-400} & \textbf{Kinetics-600}\\
\hline
\hline
Two Stream~(spatial stream)~\cite{twostream}	&Imagenet &	VGG-M		& 73 & 40.5 &-	\\ \hline 
\hline 
RGB-I3D \cite{i3d} & Imagenet & Inception v1	& 84.5 & 49.8&-	\\ 
\hline 
\hline 
\hline 
C3D \cite{c3d} & Sport1M & VGG11	& 82.3 & 51.6&-	\\ 
\hline 
TSN \cite{tsn} & Imagenet,Kinetics & Inception v3	& 93.2 & -&	72.5\\ 
\hline 
RGB-I3D \cite{i3d} & Imagenet,Kinetics & Inception v1	& 95.6 & 74.8& 72.1	\\ 
\hline 
\hline
3D ResNext 101 (16 frames) \cite{3DResHara} &Kinetics &	ResNext101	& 90.7 & 63.8&65.1	\\ 
\hline 
STC-ResNext 101 (64 frames) \cite{stcnn} &Kinetics &	ResNext101	& {96.5}		&  {74.9}& 68.7	\\ 
\hline 
ARTNet \cite{artnet} & Kinetics & ResNet18	& 93.5 & 67.6& 69.2	\\ 
\hline 
R(2+1)D \cite{R2+1} & Kinetics & ResNet50	& 96.8 & 74.5& 72	\\ 
\hline 
ir-CSN-101 \cite{Duiccv19} & Kinetics & ResNet101	& - & -& 76.7	\\ 
\hline 
DynamoNet \cite{dynamonet} & Kinetics & ResNet101	& - & -& 76.8	\\ 
\hline 
SlowFast 4$\times$16~\cite{slowfast} & Kinetics & ResNet50	& - & -& 75.6	& 78.8\\ 
\hline 
SlowFast 16$\times$8\textbf{*}~\cite{slowfast} & Kinetics & ResNet101	& - & -& 78.9\textbf{*} & 81.1	\\ 
\hline 
\textbf{HATNet~(32 frames)} & Kinetics & ResNet50 & {96.8} 	& {74.8}& 77.2 & 80.2\\ 
\hline 
\hline
\textbf{HATNet~(32 frames)} & HVU &	ResNet18 & {96.9} 	& {74.5}& {74.2} & 77.4\\ 
\hline 
\textbf{HATNet~(16 frames)} & HVU &	ResNet50 & {96.5} 	& {73.4}& {76.3} & 79.4\\ 
\hline 
\textbf{HATNet~(32 frames)} & HVU &	ResNet50 & \textbf{97.8} 	& \textbf{76.5}& \textbf{79.3}& \textbf{81.6} \\ 
\bottomrule
\end{tabular}}
\caption{State-of-the-art performance comparison on UCF101, HMDB51 test sets and Kinetics validation set. The results on UCF101 and HMDB51 are average mAP over three splits, and for Kinetics(400,600) is Top-1 mAP on validation set. For a fair comparison, here we report the performance of methods which utilize only RGB frames as input. \textbf{*}SlowFast uses multiple branches of 3D-ResNet with bigger backbones.}
\label{table:3datasets}
\vspace{-0.7cm}
\end{table*}
\section{Experiments}
In this section, we demonstrate the importance of HVU on three  different tasks: video classification, video captioning and video clustering. First, we introduce the implementation details and then show the results of each mentioned method on multi-label video recognition. Following, we compare the transfer learning ability of HVU against Kinetics. Next, as an additional experiment, we show the importance of having more categories of tags such as scenes and objects for video classification. Finally, we show the generalisation capability of HVU for video captioning and clustering tasks.  For each task, we test and compare our method with the state-of-the-art on benchmark datasets. For all experiments, we use RGB frames as input to the ConvNet. For training, we use 16 or 32 frames long video clips as single input. We use PyTorch framework for implementation and all the networks are trained on a machine with 8 V100 NVIDIA GPUs.

\subsection{HVU Results}
In Table~\ref{table:baselines}, we report the overall performance of different simpler or multi-task learning baselines and  HATNet on the HVU validation set. The reported performance is mean average precision on all of the labels/tags. HATNet that exploits both appearance and temporal information in the same pipeline achieves the best performance, since recognizing objects, scenes and attributes need an appearance module which other baselines do not have. With HATNet,  we show that combining the 3D~(temporal) and 2D~(appearance) convolutional blocks one can learn a more robust reasoning ability.

\subsection{Multi-Task Learning on HVU}
Since the HVU is a multi-task classification dataset, it is interesting to compare the performance of different deep neural networks in the multi-task learning paradigm as well. For this, we have used the same architecture as in the previous experiment, but with  different last layer of convolutions to observe multi-task learning performance. We have targeted six tasks: scene, object, action, event, attribute, and concept classification. In Table~\ref{table:multitask}, we have compared standard training without multi-task learning heads versus multi-task learning networks.

The simple baseline multi-task learning methods achieve higher performance on individual tasks as expected, in comparison to standard networks learning for all categories as a single task. Therefore this initial result on a real-world multi-task video dataset motivates the investigation of more efficient multi-task learning methods 
for video classification. 


\subsection{Transfer Learning: HVU vs Kinetics}
Here, we study the ability of transfer learning with the HVU dataset. We compare the results of pre-training  3D-ResNet18 using Kinetics versus using HVU and then fine-tuning on UCF101, HMDB51 and Kinetics. Obviously, there is a large benefit from pre-training of deep 3D-ConvNets and then fine-tune them on smaller datasets (i.e. HVU, Kinetics $\Rightarrow$ UCF101 and HMDB51). As it can be observed in Table~\ref{table:TFlearn}, models pre-trained on our HVU dataset performed notably better than models pre-trained on the Kinetics dataset. Moreover, pre-training on HVU can improve the results on Kinetics also.

\subsection{Benefit of Multiple Semantic Categories}
Here, we study the effect of training models with multiple semantic categories, in comparison to using only a single semantic category, such as Kinetics which covers only action category. In particular, we designed an experiment by having the model trained in multiple steps by adding different categories of tags one by one. Specifically, we first train 3D-ResNet18 with action tags of HVU, following in second step we add tags from object category and in the last step we add tags from the scene category. For performance evaluation, we consider action category of HVU. In the first step the gained performance was 43.6\% accuracy and after second step it was improved to 44.5\% and finally in the last step it raised to 45.6\%. The results show that adding high-level categories to the training, boosts the performance for action recognition in each step. As it was also shown in Table~\ref{table:multitask}, training all the categories together yields 47.5\% for the action category which is $\sim$4\% gain over action as single category for training. Thus we can infer from this that an effective feature representation can be learned by adding additional categories, and also acquire knowledge for an in-depth understanding of the video in holistic sense. \vspace{-0.2cm}

\begin{table}[t] 
\centering
\small
\tabcolsep=0.10cm
\resizebox{9cm}{!}{
\begin{tabular}{ c|c|  c }
\toprule
Model& Pre-Training Dataset & BLEU@4 \\
\hline 
\hline 
SA(VGG+C3D)~\cite{MSRVTT} & ImageNet+Sports1M &  36.6 \\
M3(VGG+C3D)~\cite{m3} & ImageNet+Sports1M  &  38.1 \\
SibNet(GoogleNet)~\cite{sibnet} & ImageNet &  40.9 \\
MGSA(Inception+C3D)~\cite{mgsa} & ImageNet+Sports1M &  42.4 \\
\hline
I3D+M~\cite{xgate} & Kinetics &  41.7\\
3D-ResNet50+M & Kinetics &  41.8\\
3D-ResNet50+M & HVU &  \textbf{42.7}\\
\bottomrule
\end{tabular}}
\caption{Captioning performance comparisons of \cite{xgate} with different models and pre-training datasets. M denotes the motion features from optical flow extracted as in the original paper.} \label{table:captioning}
\vspace{-0.8cm}
\end{table}

\subsection{Comparison on UCF, HMDB, Kinetics}
In Table~\ref{table:3datasets}, we compare the HATNet performance with the state-of-the-art on UCF101, HMDB51 and Kinetics. For our baselines and HATNet, we employ pre-training in two separate setups: one with HVU and another with Kinetics, and then fine-tune on the target datasets. For UCF101 and HMDB51, we report the average accuracy over all three splits. We have used ResNet18/50 as backbone model for all of our networks with 16 and 32 input-frames. HATNet pre-trained on HVU with 32 frames input achieved superior performance on all three datasets with standard network backbones. 
Note that on Kinetics, HATNet even with ResNet18 as a backbone ConvNet performs almost comparable to SlowFast which is trained by dual 3D-ResNet50. In Table~\ref{table:3datasets}, however while SlowFast has better performance using dual 3D-ResNet101 architecture, HATNet obtains comparable results with much smaller backbone.\vspace{-0.4cm}

\subsection{Video Captioning}
We present a second task that showcases the effectiveness of our HVU dataset, we evaluate the effectiveness of HVU for video captioning task. We conduct experiments on a large-scale video captioning dataset, namely MSR-VTT~\cite{MSRVTT}. We follow the standard training/testing splits and protocols provided originally in~\cite{MSRVTT}. For video captioning, the performance is measured using the BLEU metric.  

\noindent\textbf{Method and Results:} Most of the state-of-the-art video captioning methods use models pre-trained on Kinetics or other video recognition datasets. With this experiment, we intend to show another generalisation capability of HVU dataset where we evaluate the performance of pre-trained models trained on HVU against Kinetics. For our experiment, we use the Controllable Gated Network~\cite{xgate} method, which is to the best of our knowledge the state-of-the-art for captioning task. 

For comparison, we considered two models of 3D-ResNet50, pre-trained on (i) Kinetics and (ii) HVU. Table~\ref{table:captioning} shows that the model trained on HVU obtained better gains in comparison to Kinetics. This shows HVU helps to learn more generic video representation to achieve better performance in other tasks.
\vspace{-0.2cm}



\subsection{Video Clustering}
With this experiment, we evaluate the effectiveness of generic features learned using HVU when compared to Kinetics.  

\noindent\textbf{Dataset:} We conduct experiments on ActivityNet-100~\cite{actnet} dataset. For this experiment  we provide results when considering 20 action categories with 1500 test videos. We have selected ActivityNet dataset to make sure there are no same videos in HVU and Kinetics training set. For clustering, the performance is measured using clustering accuracy~\cite{sharmafg}. 

\noindent\textbf{Method and Results:} We extract features using 3D-ResNet50 and HATNet pre-trained on Kinetics-600 and HVU for the test videos and then cluster them with KMeans clustering algorithm with the given number of action categories. Table~\ref{table:cluster} clearly shows that the features learned using HVU is far more effective compared to features learned using Kinetics.
\vspace{-0.2cm}

\begin{table}[t] 
\centering
\small
\tabcolsep=0.15cm
\resizebox{9cm}{!}{
\begin{tabular}{ c|c| c }
\toprule
Model& Pre-Training Dataset & Clustering Accuracy~(\%) \\
\hline 
\hline 
3D-ResNet50 & Kinetics & 50.3\\
3D-ResNet50 & HVU & 53.5\\
HATNet & HVU & \textbf{54.8}\\ 
\bottomrule
\end{tabular}}
\caption{Video clustering performance: evaluation based on extracted features from networks pre-trained on Kinetics and HVU datasets.} \label{table:cluster}
\vspace{-0.8cm}
\end{table}